\documentclass{article}

\usepackage{microtype}
\usepackage{graphicx}
\usepackage{booktabs} 

\usepackage[rejected]{mlsys2023}

\usepackage{natbib}
\PassOptionsToPackage{sort}{natbib}
\usepackage[utf8]{inputenc} 
\usepackage[T1]{fontenc}    
\usepackage[hyphens]{url}   
\usepackage{hyperref}
\usepackage{booktabs}       
\usepackage{amsfonts}       
\usepackage{amssymb}        
\usepackage{nicefrac}       
\usepackage{microtype}      
\usepackage{color}
\usepackage{listings}
\usepackage{comment}
\usepackage{mathtools}      
\usepackage{caption}
\usepackage{subcaption}
\usepackage[export]{adjustbox}
\definecolor{commentgreen}{RGB}{2,200,10}
\definecolor{weborange}{RGB}{255,165,0}
\definecolor{purple}{RGB}{159,90,253}
\definecolor{comment}{RGB}{106,153,85}
\definecolor{darkteal}{RGB}{4,93,93}

\usepackage{inconsolata}
\usepackage{graphicx}

\usepackage{ifthen}

\newcommand{\?}{\stackrel{?}{=}}

\newcommand{\final}{1}

\newcommand{\todo}   [1]{{{\color{darkteal} #1}}}
\ifthenelse{\equal{\final}{1}}{\renewcommand{\todo}[1]{}}{}

\newcommand{\mosama}   [1]{{{\color{purple}(mosama) #1}}}
\ifthenelse{\equal{\final}{1}}{\renewcommand{\mosama}[1]{}}{}

\newcommand{\collin}   [1]{{{\color{weborange}(collin) #1}}}
\ifthenelse{\equal{\final}{1}}{\renewcommand{\collin}[1]{}}{}

\newcommand{\jowens}   [1]{{{\color{blue}(jowens) #1}}}
\ifthenelse{\equal{\final}{1}}{\renewcommand{\jowens}[1]{}}{}

\newcommand{\jowensps}   [1]{{{\color{blue}(jowens post-submission) #1}}}
\ifthenelse{\equal{\final}{1}}{\renewcommand{\jowensps}[1]{}}{}

\newcommand{\cameron}   [1]{{{\color{magenta}(cameron) #1}}}
\ifthenelse{\equal{\final}{1}}{\renewcommand{\cameron}[1]{}}{}

\newcommand{\saurav}   [1]{{{\color{commentgreen}(saurav) #1}}}
\ifthenelse{\equal{\final}{1}}{\renewcommand{\saurav}[1]{}}{}

\usepackage{array}

\newcommand{\dnnz}{\ensuremath{\textit{nnz}}}

\mlsystitlerunning{The Sparsity Roofline: Understanding the Hardware Limits of Sparse Neural Networks}
 
\begin{document}

\twocolumn[
\mlsystitle{The Sparsity Roofline: Understanding the Hardware Limits of Sparse Neural Networks}



\mlsyssetsymbol{equal}{*}

\begin{mlsysauthorlist}
\mlsysauthor{Cameron Shinn}{equal,ucd}
\mlsysauthor{Collin McCarthy}{equal,ucd}
\mlsysauthor{Saurav Muralidharan}{nv}
\mlsysauthor{Muhammad Osama}{ucd}
\mlsysauthor{John D. Owens}{ucd}
\end{mlsysauthorlist}

\mlsysaffiliation{ucd}{Electrical and Computer Engineering, UC Davis, Davis, CA, United States}  
\mlsysaffiliation{nv}{NVIDIA, Santa Clara, CA, United States}

\mlsyscorrespondingauthor{Cameron Shinn}{ctshinn@ucdavis.edu}

\mlsyskeywords{Machine Learning, MLSys}

\vskip 0.3in

\begin{abstract}
    We introduce the Sparsity Roofline, a visual performance model for evaluating sparsity in neural networks. The Sparsity Roofline jointly models network accuracy, sparsity, and theoretical inference speedup. Our approach does not require implementing and benchmarking optimized kernels, and the theoretical speedup becomes equal to the actual speedup when the corresponding dense and sparse kernels are well-optimized. We achieve this through a novel analytical model for predicting sparse network performance, and validate the predicted speedup using several real-world computer vision architectures pruned across a range of sparsity patterns and degrees. We demonstrate the utility and ease-of-use of our model through two case studies: (1) we show how machine learning researchers can predict the performance of unimplemented or unoptimized block-structured sparsity patterns, and (2) we show how hardware designers can predict the performance implications of new sparsity patterns and sparse data formats in hardware. In both scenarios, the Sparsity Roofline helps performance experts identify sparsity regimes with the highest performance potential.



\end{abstract}

]



\printAffiliationsAndNotice{\mlsysEqualContribution} 

\section{Introduction}
\label{sec:introduction}

Deep neural networks are often over-parameterized~\citep{Howard:2019:SFM, Tan:2019:ERM} and their weights or parameters can be eliminated (\emph{pruned}) to improve inference latency and/or decrease network size~\cite{LeCun:1989:OBD, Han:2015:LBW, Molchanov:2017:VDS, Zhu:2018:TPO} without affecting accuracy.
Depending on the \emph{pattern} and \emph{degree} of sparsity, which together constitute a \emph{sparsity configuration}, networks exhibit widely different accuracy and runtime behavior.
This presents major problems for machine learning practitioners who wish to find the best sparsity pattern and degree that balances accuracy loss and performance constraints for their specific application.
Obtaining the accuracy corresponding to a sparsity pattern and degree typically requires some form of network fine-tuning~\cite{Frankle:2019:TLT}, making it highly inefficient to estimate the impact of different sparsity configurations by trying hundreds of combinations of hyperparameters.

Thus we hope to predict which sparsity combinations might be most fruitful without fine-tuning them all. But accurately estimating the effects that a specific sparsity configuration has on inference runtime poses a different set of challenges: (1) which metric should we use to estimate runtime performance, and (2) how do we obtain the runtime performance of sparsity patterns that are either unimplemented or have unoptimized implementations?
To illustrate the challenge of identifying the right metric, consider the total floating point operations (FLOPs) performed during sparse matrix operations such as matrix multiplication (a common operation in neural networks~\cite{Chetlur:2014:CEP}).
FLOPs are frequently used to evaluate the performance of pruned models~\cite{Han:2015:LBW, Molchanov:2017:VDS, Zhu:2018:TPO, Frankle:2019:TLT, Lee:2019:SSN, Hoefler:2021:SID, Blalock:2020:WIT}.
Table~\ref{tab:structure_vs_unstructure} illustrates the limitations of this metric. Here, we show two weight matrices that provide a counterexample to the notion that FLOPs are positively correlated with measured runtime.
The structured weight matrix shown on the left side of the table has 1.57$\times$ more FLOPs than the unstructured matrix on the right, but runs nearly 6$\times$ faster.

Addressing the challenge of \emph{estimating} optimized runtime performance is even harder.
While performance experts have implemented computation kernels specifically targeting sparse neural networks~\cite{Gale:2020:SGK, Wang:2020:SAU, Zhaodong:2021:ETC, Vooturi:2019:ESN}, there are significant gaps.
For example, NVIDIA's cuSparse library provides optimized GPU kernels for block-sparse matrices, but they are primarily optimized for larger block sizes such as 16$\times$16 and 32$\times$32~\cite{Yamaguchi:2021:AMM}.

As discussed in Section~\ref{sec:case-study-dl-practitioner}, using smaller block sizes often leads to higher accuracies; however, in the absence of computation kernels optimized for these sizes, it is impossible to estimate their effect on runtime via benchmarking.

\begin{table}[]
  \caption{\textbf{Runtime vs. GFLOPs}: SpMM performance on (32$\times$32) block sparsity vs.\ unstructured with a similar amount of nonzeros. White indicates zero-valued weights, blue non-zero. The block sparse matrix has more FLOPs but has a nearly 6$\times$ better runtime latency vs.\ unstructured.} 
  \label{tab:structure_vs_unstructure}
  \centering
  \resizebox{\columnwidth}{!}{%
  \begin{tabular}{@{}rcc@{}}
    \toprule
    & \begin{tabular}[c]{@{}c@{}}Structured \\ (32$\times$32)\end{tabular} & Unstructured \\
    \midrule
    Matrix Heatmap &
    \includegraphics[height=0.5\linewidth,valign=b]{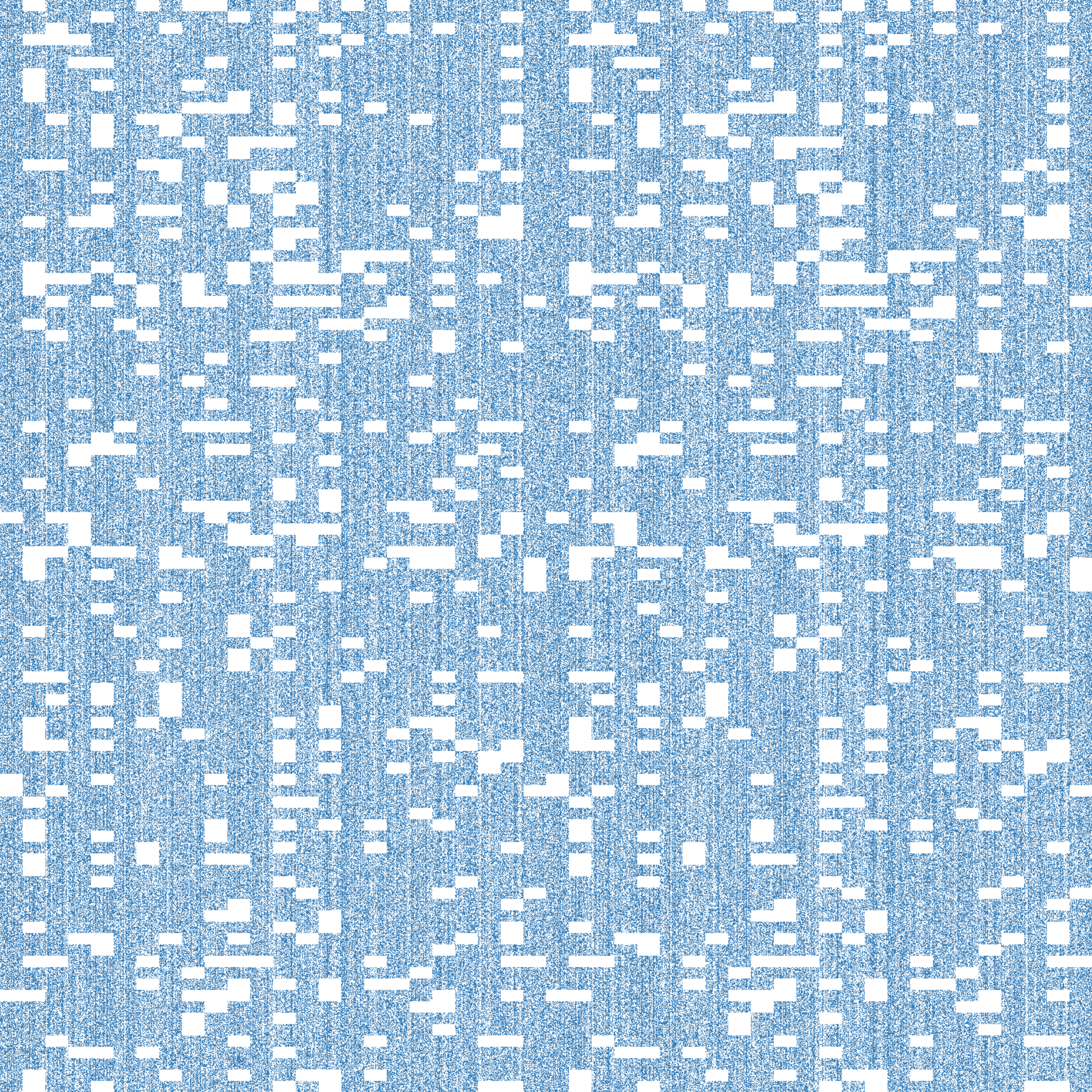} & \includegraphics[height=0.5\linewidth,valign=b]{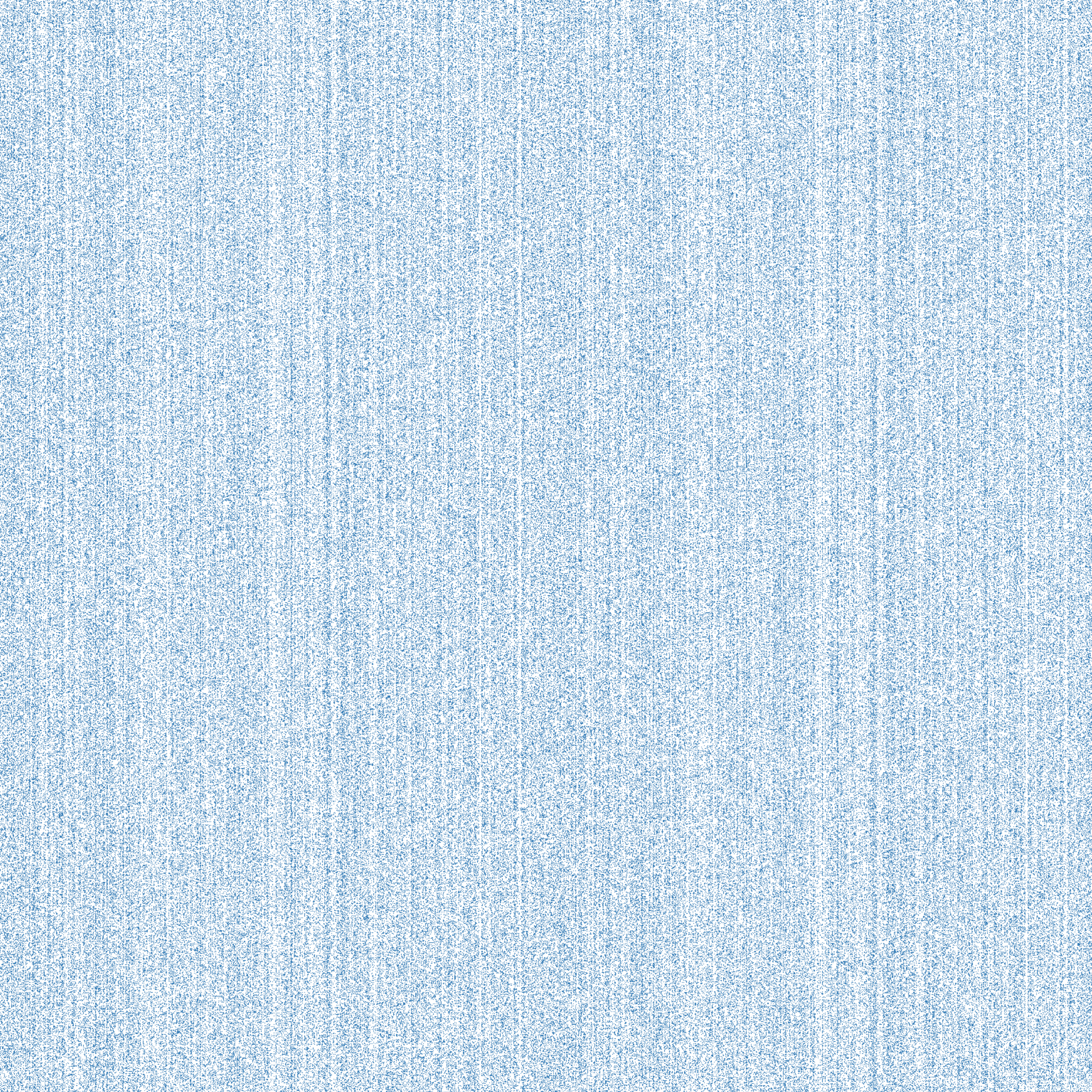} \\
    & & \\
    \midrule
    \textbf{Runtime (ms)} & \textbf{0.613} & \textbf{3.526} \\
    \textbf{GFLOPs} & \textbf{24.4} & \textbf{15.5} \\
    TFLOPs/s & 39.9 & 4.4 \\
    Number of Nonzeros & 1.95M & 1.23M \\
    \begin{tabular}[x]{@{}r@{}}$m\text{-} \times k\text{-}$dimensions\\\footnotesize{(sparse operand)}\end{tabular} & 3072$\times$768 & 3072$\times$768 \\
    \begin{tabular}[x]{@{}r@{}}$n$-dimension\\\footnotesize{(dense operand)}\end{tabular} & 6272 & 6272 \\
    \bottomrule
  \end{tabular}
  } %
\end{table}

To help practitioners better understand the complex relationship between sparsity configuration, accuracy, and inference performance (both current and potential), we introduce a novel visual model named the \emph{Sparsity Roofline}.
Our work builds upon the well-known Roofline model~\cite{Williams:2009:RAI}, which provides a visual representation of the performance of a given computation kernel.

In the Roofline model, users compute the \emph{arithmetic intensity} of the given kernel, and plot it against one or more hardware-specific upper limits (the Rooflines) defined by the peak memory bandwidth and peak floating-point throughput of that hardware architecture.
In a similar vein, the Sparsity Roofline plots network accuracy against the theoretical speedup of sparse over dense models, with additional sparsity information. \jowens{This last phrase is vague. Instead ``given the  characteristics of the target sparsity configuration'' or something like that?}
This clearly shows the two most important aspects of weight pruning to a machine learning practitioner---accuracy and performance---and can be analyzed across any model architecture, sparsity hyperparameters, or hardware accelerator.
Plotting the Sparsity Roofline requires sampling the accuracy values corresponding to the sparsity configurations being analyzed, which can be easily done with masking-based approaches and existing software libraries~\cite{Paszke:2019:PAI, Joseph:2020:APA}.
The only other metrics needed are \mosama{The previous part of the sentence reads like ``there's more that you need'' (negatively), instead ``bridge'' the ML-specific metrics with problem/hardware specific metrics by collecting\dots} the arithmetic intensity\saurav{of what?}\jowens{$\leftarrow$ what Saurav said}, which can be either profiled or computed \mosama{estimated?} by hand, and the hardware-specific peak computational throughput (in FLOPs/s) and memory bandwidth (in bytes/s).
\saurav{Highlighted text above is missing details: explain how accuracy is obtained for multiple points (novel sampling strategy).}

We validate and demonstrate the usefulness of the Sparsity Roofline by analyzing several real-world computer vision models, including convolutional neural networks (CNNs), vision transformers (ViT), and multi-layer perceptron (MLP)-based networks.
We investigate which sparsity characteristics have the greatest impact on accuracy and GPU performance, and point out promising areas to focus on for kernel optimization.
Finally, we present two case studies: (1) analyzing tradeoffs associated with block-structured sparsity for deep learning practitioners, and (2) efficient sparsity patterns for future hardware architectures.
\saurav{Revisit claims in highlighted text after validation section is written.}

This paper makes the following contributions:

\begin{enumerate}
    \item
      It introduces the Sparsity Roofline visual model for understanding accuracy vs.\ latency trade-offs for currently unoptimized and unimplemented kernel designs.
    \item
      It uses the Sparsity Roofline to benchmark and analyze several real-world computer vision architectures pruned to a range of sparsity patterns and levels.
    \item
      It demonstrates the use of the Sparsity Roofline in two distinct use cases: to analyze block-sparsity structures for DL practitioners, and to help inform future sparse hardware implementations.
\end{enumerate}

\section{Background}
\label{sec:background}



In this Section, we provide a brief overview of neural network pruning, followed by a description of the traditional Roofline model.

\subsection{Neural Network Pruning} Weight pruning involves setting a subset of neural network weights to zero, followed by a training or fine-tuning stage that attempts to recover any lost accuracy~\cite{Hoefler:2021:SID}.
Pruning can be unstructured (fine-grained), where individual non-zero values are eliminated, or structured (coarse-grained), where groups of non-zero values are removed instead, each resulting in a different \emph{sparsity pattern}.
The \emph{sparsity level} refers to the fraction of zero weights to total weights and is expressed as a percentage in this paper.
Structured pruning has been demonstrated to achieve better runtime performance, typically at the cost of decreased accuracy~\cite{Narang:2017:BSR, Vooturi:2018:HBS, Li:2022:RRC}.
A number of algorithms have been proposed in the literature for accuracy recovery of pruned models~\cite{Deng:2021:CSC, Renda:2020:CRF, Hoefler:2021:SID}. In this paper, we use the learning rate rewinding approach proposed by \citet{Renda:2020:CRF}.

\jowens{Be consistent: ``tensor cores'', ``Tensor cores'', or ``Tensor Cores'' are all used in this paper. I would probably recommend just the simplest ``tensor cores''.}

\subsection{The Roofline Model} The Roofline model~\cite{Williams:2009:RAI} is a visual performance model that shows how well a computational kernel utilizes the hardware. The Roofline model plots the arithmetic intensity (FLOPs computed / bytes read and written) on the x-axis and the throughput (FLOPs per second) on the y-axis. This enables users to visually observe if their program is memory-bound or compute-bound, and to what extent. The upper bound (Roofline) of the model is determined by both the hardware's peak compute throughput and peak memory bandwidth. Although there are variants that consider cache hierarchies~\cite{Ilic:2014:CAR}, the traditional Roofline model that we discuss in this paper assumes perfect caching is necessary (including user-managed caching such as shared memory and local registers) to achieve peak memory bandwidth utilization; we thus use DRAM memory bandwidth. The hardware throughput component can be increased with additional hardware acceleration for a specific application (e.g., Tensor Cores for deep learning~\cite{Jia:2018:DTN}). The utility of the Roofline model comes from its ability to succinctly show potential improvement for a given program with respect to the hardware speed-of-light.

\subsection{Evaluating Sparse Neural Networks}
\label{sec:evaluating-sparse-models}

\begin{figure}[t]
    \centering
    \includegraphics[width=\linewidth,valign=t]{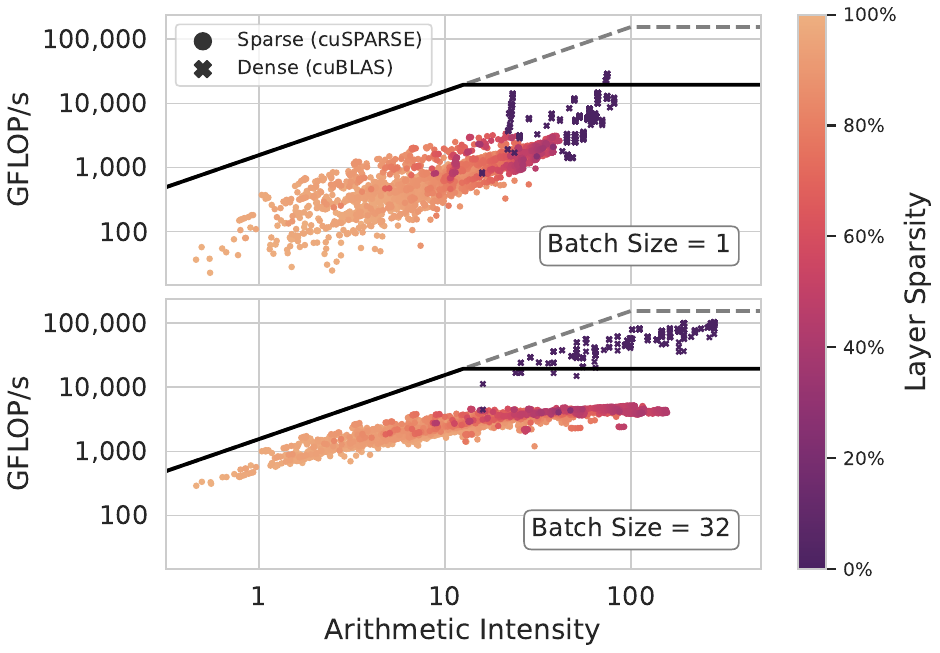}
    \caption{\textbf{Roofline, Sparse vs. Dense}: Roofline model measuring throughput of SpMM on unstructured sparse layers and GEMM on dense layers from all \jowens{replace ``all'' with a number} trained models, on a single NVIDIA A100. The solid line is the CUDA core peak throughput, the dashed line the Tensor core peak throughput. Unstructured sparsity kernels in cuSPARSE do not use Tensor cores.\collin{Should we call this ``operational intensity'' or should we add ``arithmetic intensity'' to related works when we introduce the roofline?}  \mosama{The caption explains nicely what I am looking at but what conclusion should I draw from a first glance? It could be as general as ``We can visualize the potential performance speedup\dots''} \jowens{AI should have units. I think properly the $y$ axis should be ``GFLOPs/s''.}}
    \label{fig:stacked_unstr_spmm_roofline}
\end{figure}

Figure~\ref{fig:stacked_unstr_spmm_roofline} plots the Roofline for individual SpMM matrices across all \jowens{``all''? I think I'd just use a number here, because the reader will say ``all? what is all?''. You can explain later in the evaluation section (and use the same number there).} benchmarked computer vision models. The line in each plot is the ``Roofline'', which slopes upwards during the memory-bound region where the arithmetic intensity (AI) is too low to saturate the compute resources, and flattens out once the AI reaches a hardware-specific point, called the \emph{knee}. The dashed line is for Tensor Cores and the solid line for CUDA cores, where the Tensor Core knee has almost 10x the AI of CUDA cores.

The points that are closest to the Roofline are utilizing the GPU the best, with higher sparsities being more memory bound and lower sparsities approaching and becoming compute bound in some situations, such as when the inner-dimension of the matrix product is higher. \jowens{Because it shows \ldots,} The Roofline model is a significant improvement over analyzing FLOPs, but it has three major drawbacks in optimizing sparse deep learning models:

\collin{Maybe the itemize isn't the best, now that I'm looking at the PDF\@. Leaving it for now.}

\begin{enumerate}
    \item The Roofline model lacks any concept of accuracy, and GFLOPs/s is challenging to use to compare the relative performance between sparse and dense layers.
    \item The Roofline model is only meaningful per-layer instead of per-model. An entire model is almost always a combination of layers, where some are memory-bound and others are likely compute-bound. Therefore calling the entire model ``compute bound'' or ``memory bound'' is misleading at best.
    \item The Roofline model requires benchmarking to compute GFLOPs/s. Even if optimal kernels exist, such as cuBLAS for dense GEMM operations, the surrounding benchmarking framework is time-consuming to implement, test, and maintain.
\end{enumerate}

Our proposed solution, the Sparsity Roofline, directly addresses these concerns. It is not meant to replace the Roofline model, but instead \emph{complement} it for the specific use case of designing and optimizing sparse deep-learning kernels.




\section{The Sparsity Roofline}
\label{sec:sparse-roofline}

\begin{figure}[tb]
    \centering
    \includegraphics[width=\linewidth,valign=t]{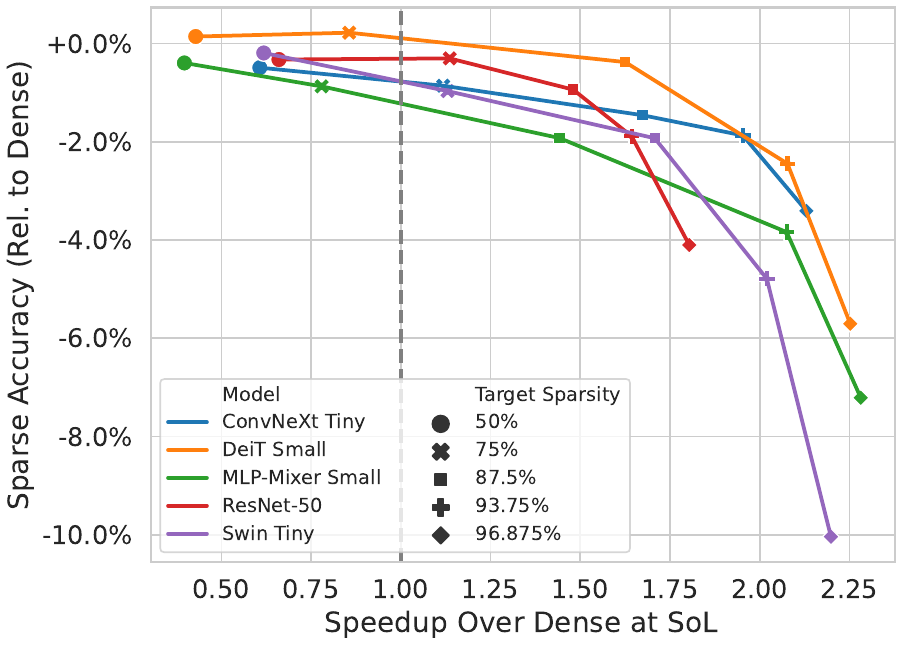}
    \caption{\textbf{Per-Model Sparsity Roofline}: The Sparsity Roofline for several computer vision models on ImageNet-100 pruned with global magnitude pruning. Speedup is calculated per-layer using the maximum compute or memory bound latency, and then summed per model. The machine learning engineer can choose the architecture that provides the optimal balance of accuracy, speedup, and implementation difficulty. \mosama{Feels weird that this figure comes \emph{before} the Table~\ref{tab:structure_vs_unstructure}. We reference the table earlier in text.}}
    \label{fig:unstr_sparsity_roofline_b1}
  \end{figure}

The Sparsity Roofline is designed to be an easy-to-use tool for deep learning practitioners interested in sparsity, performance experts, and hardware designers. It achieves this goal by addressing the three major issues with the existing Roofline model described in Section~\ref{sec:evaluating-sparse-models}. 

\jowens{My take on the s.r.\ is that at least one major benefit of the visualization is the ability to compare different sparsity regimes. That might not be the point you're trying to make, and that's OK, just ignore what I'm saying then, but if that is important, that is not clear enough to me. I will suggest an edit below.}

The Sparsity Roofline plots accuracy vs.\ theoretical speedup, as opposed to the traditional Roofline's GFLOPs/s vs.\ arithmetic intensity. Accuracy is almost always the most important optimization metric in DNNs, and therefore we place it on the $y$ axis. Similarly, replacing GFLOPs/s with theoretical speedup makes it far easier to understand relative performance differences of a sparse and dense layer or model. Further, the sparsity configuration \jowens{is a first-class entity in the Sparsity Roofline, with different sparsity configurations easily comparable via \ldots (delete ``is encoded into'' next)} is encoded into the point and/or line style in order to easily compare different sparsity design decisions, which are crucial for optimal performance.

The Sparsity Roofline converts per-layer peak GFLOPs/s to per-model minimum or \emph{speed-of-light} (SoL) latency. We first calculate a per-layer SoL latency, then sum the layer-wise latencies for the model SoL latency. This represents the true performance metric that practitioners care about: end-to-end latency of the entire model.

Like the traditional Roofline, the Sparsity Roofline does not require benchmarking. We only need to look up the hardware peak GFLOPs/s and peak GB/s of a hardware architecture, and compute the per-layer GFLOPs and GBs read/written by hand in order to calculate arithmetic intensity.
    \saurav{Not clear how this paragraph contrasts Sparsity Roofline with traditional Roofline. Neither require benchmarking.} \jowens{I edited to address this point. It's OK with me if this point does not contrast R and SR\@. It's an important point to make even if it's not different than R\@.}

The Sparsity Roofline for unstructured sparsity is shown in Figure~\ref{fig:unstr_sparsity_roofline_b1}, and for ConvNeXt-Tiny and Swin-Tiny in Figures~\ref{fig:block_sparsity_roofline} and~\ref{fig:nofm_sparsity_roofline}, respectively. We will now describe how these Sparsity Rooflines are constructed.

Given our model uses accuracy metrics, the model being used in the Sparsity Roofline needs to be fine-tuned to a given sparsity from a pre-trained dense model. Fine-tuning for sparsification is a standard practice in deep learning, and the only way to quantify accuracy. We use the learning-rate rewinding technique proposed by \citet{Renda:2020:CRF} and the Condensa library by \citet{Joseph:2020:APA}.
Our model is most accurate when the sparse kernels are well optimized, and thus approaching the speed-of-light. This doesn't mean the sparse kernel needs to be compute bound, but if it is memory bound the closer it is to the device peak memory throughput, the more accurate our model is. This is discussed in detail in Section~\ref{sec:validation}.
\saurav{Move this elsewhere.}



\subsection{Use Cases}

\collin{Some of this may be repetitive and not strictly necessary, or it may just need to be refactored. I'm trying to make sure we include $\sim$everything we had in bullet-points previously, including John's comments, for the first pass at least.}

\saurav{I vote for removing this subsection entirely.}

The Sparsity Roofline is designed to quantify the performance-accuracy tradeoff for a specific combination of hardware, model architecture and sparsity configuration, such as sparsity pattern, sparsity level or percent, and sparse data format. Thus it can be used by both software and hardware engineers who want to understand how an optimized kernel would perform, but do not want to go through the trouble of implementing and benchmarking sub-optimal scenarios. In Section~\ref{sec:case-study-dl-practitioner}, we show how a deep-learning practitioner may use this tool to investigate optimal block-structure sparsity patterns, and in Section~\ref{sec:case-study-hardware-architect} we show how a hardware engineer  can investigate different N:M sparsity patterns and sparse data formats to implement in hardware, e.g., for new sparse Tensor core formats.

In contrast, the Sparsity Roofline is not meant for engineers who already have a specific sparsity-configuration optimization target. In that scenario, a combination of the Roofline model, benchmarking / profiling, and lower-level optimizations are likely the correct tools to understand detailed performance statistics that would inform kernel design, such as load balancing and caching. 

\subsection{Constructing the Sparsity Roofline}

\collin{We should reduce some vertical whitespace here, between the tops of equations and the text, not sure what the preferred way to do that is.} \jowens{I added two setlength calls below; they're set right now to save some space over the default; feel free to adjust, even to make them negative.} \collin{Also I've removed the equation numbers since we're not using them, and the text is long enough that the number is getting wrapped to the next line which looks weird. It's possible we should keep the equation numbers anyway, per requirements or general style (not sure).}

The Sparsity Roofline plots accuracy vs.\ theoretical speedup from sparsity. We start by deriving the theoretical speedup.

First, we need to define the kernel's GFLOPs and GBs read/written to global memory. Equation~\ref{eq:spmm_flops_gb} shows this for SpMM ($\text{Sparse} \times \text{Dense} = \text{Dense}$ matrix multiply); the index data depends on the sparse data format. For compressed row format (CSR), it is $\dnnz + m + 1$.

\setlength{\abovedisplayskip}{0pt}
\setlength{\belowdisplayskip}{0pt}

\begin{equation}
\begin{aligned} \label{eq:spmm_flops_gb}
    \text{SpMM FLOPs} &= \dnnz \times n \\
    \text{SpMM GB} &= \dnnz + n \times k + m \times n + \text{index data}
\end{aligned}
\end{equation}

Next, we define the per-layer speed-of-light latency as the maximum runtime for the kernel's given GFLOPs and GBs read/written to global memory. Using the device's peak GFLOPs and GB/s, this is computed as


\begin{equation} \label{eq:sol_runtime2}
    \text{Per-Layer SoL} = \text{max}\bigg(
        \frac{\text{GFLOP}}{\text{Peak GFLOP/s}}, \frac{\text{GB}}{\text{Peak GB/s}}
        \bigg)
\end{equation}
\collin{Is that comma the correct? Looks weird. Previous horizontal-version is commented out above.} \jowens{yes, it's correct, it's lined up with the baseline, just leave it as is.}

Finally, we sum the $L$ per-layer runtimes for the dense model and the same corresponding sparse model, and take their runtime ratio as the speedup, using the dense computation as the baseline. For example, if the sparse latency is 1~ms and the dense latency is 2~ms, the speedup would be 2x. \collin{This still sounds / feels weird (as in I want to just define it as 0.5, so flipped), but I think this is a correct way to say it.}

\begin{equation} \label{eq:speedup_at_sol}
    \text{Speedup at SoL} = \\
    \frac{\sum_{l = 1}^{L_{\text{dense}}} \text{SoL Runtime}_l}{\sum_{l = 1}^{L_{\text{sparse}}} \text{SoL Runtime}_l}
\end{equation}

These equations make the same assumption as the Roofline model: the maximum achievable FLOPs/s is the hardware's peak compute throughput, and each byte of data may be read from or written to global memory once, at the hardware's peak memory throughput, with perfect caching (including shared memory or local registers) for any intermediate reads.

\subsection{Evaluating Accuracy}

\jowens{It is not particularly clear to me what the procedure is for SR\@. There's work to be done per architecture, per model, and per sparsity configuration? There's evidently two stages, corresponding to the two paragraphs below? If I was reading this (and the previous subsection), it's not nearly clear enough as to what I need to actually do to build my own SR\@.}

To compute accuracy for each model and sparsity configuration, we start by pre-training one baseline model per architecture. \jowens{This contrasts with the assertion in the previous section that SR does not require benchmarking.} We pre-train without sparsity for 300 epochs on ImageNet-100~\cite{Vinyals:2016:MNF}. This dataset is a subset of the ImageNet-1K dataset~\cite{Deng:2009:IAL} created by sampling 100 of the 1000 classes in ImageNet-1K, which allows us to train a larger number of models, sparsity patterns, and sparsity levels. 

\saurav{The paragraph above is misleading. The user starts with their own pre-trained model of interest. We seem to be implying that this pre-training is part of constructing the S.R. Might be best to move this to the evaluation methodology.}

All model definitions are from the \emph{timm} library~\cite{Wightman:2019:PIM} and each is trained with the same set of data augmentations, hyperparameters, and training schedules based on modern architectures such as DeiT~\cite{Touvron:2021:TDE}, Swin~\cite{Liu:2021:STH} and ConvNeXt~\cite{Liu:2022:ACF}.
This includes data augmentations RandAugment~\cite{Cubuk:2020:RPA}, MixUp~\cite{Zhang:2018:MBE} and CutMix~\cite{Yun:2019:CRS}, a cosine decay learning rate schedule~\cite{Loshchilov:2017:SGD}, and the AdamW optimizer~\cite{Loshchilov:2019:DWD} with a base learning rate of $10^{-3}$ and 20 epochs of warm up.
Using these uniform settings across all models ensures a fair comparison with an identical training procedure. We store the checkpoint with the minimum validation loss and use this for fine-tuning.

We apply an incremental fine-tuning algorithm based on learning rate rewinding~\cite{Renda:2020:CRF} to the baseline model to obtain the accuracy values corresonding to the following sparsity levels: 50\%, 75\%, 87.5\%, 93.75\% and 96.875\%. This pattern involves halving the number of nonzeros per iteration, which ends up slightly biasing the results towards higher sparsities where sparse kernels are typically more performant.

For a given combination of model and sparsity pattern, e.g., ConvNeXt-Tiny with unstructured sparsity, we prune the weights with global magnitude pruning to the lowest sparsity level of 50\%. We rewind the learning rate schedule but with a shorter 60 epoch total decay rather than 300 epochs.
After 60 epochs we increase the sparsity level by $(1 - \text{Sparsity}) / 2$, prune the additional weights, and rewind the learning rate again.
We repeat this a total of five times within a single run to fine-tune five sparsity levels for our model / sparsity pattern combination in 300 epochs total, which is the same number as during training.
We find this process to be simple and efficient, and quantitatively works well for ImageNet-100. For more challenging datasets such as ImageNet-1k or ImageNet-22k, the fine-tuning schedule would likely need to be increased.







\subsection{Validation}
\label{sec:validation}

\collin{Deriving when SoL speedup == measured speedup is easy, but simplifying the absolute or relative error (SoL speedup - measured speedup) seems to be more complicated. It is probably easiest to plot using actual data if there's time. Again I'm not liking the extra white space but between \\begin\{equation*\} and the the first equation but I'm not sure what the best way to fix it is (or if we're allowed to)}

It is important to understand the cases where speed-of-light (SoL) speed-up equals the actual measured speed-up, without having to implement and optimize a specific sparse kernel. We can easily show that the speedup at SoL is precisely equal to the measured speed-up when the sparse and dense kernels are \emph{equally optimized}. Specifically, at a per-layer level this occurs when the percentage of the per-layer SoL latency for dense and sparse are equal. For example, if a given GEMM kernel is compute bound and obtains 90\% of the SoL GFLOPs/s, and the corresponding SpMM kernel is memory bound and also obtains 90\% of the SoL GB/s, then the percent of SoL is identical and our model will predict a SoL speedup that is equal to the measured speedup.
More formally:

\begin{equation*}
\begin{aligned}
  \text{Per-Layer Speedup at SoL} &\? \text{Per-Layer Speedup Meas.} \\[3pt]
  \frac{\text{Dense SoL Runtime}}{\text{Sparse SoL Runtime}} &= \frac{\text{Dense Meas. Runtime}}{\text{Sparse Meas. Runtime}} \\[3pt]
  \frac{\text{Dense SoL Runtime}}{\text{Dense Meas. Runtime}} &= \frac{\text{Sparse SoL Runtime}}{\text{Sparse Meas. Runtime}} \\[3pt]
  \text{Dense Per-Layer \% of SoL} &= \text{Sparse Per-Layer \% of SoL}
\end{aligned}
\end{equation*}

In the last equation, note that the percent of speed-of-light (or fraction of speed-of-light) is defined as the ratio between the SoL latency to the measured latency. The measured latency can take on values as small as the SoL latency but no smaller, by definition. Therefore this is bounded between 0--1 (or 0--100\%).

The same equation holds for per-model aggregation, but in this case each individual term is a summation of all layers.

\begin{equation*}
  \begin{aligned}
    \text{Per-Model Speedup at SoL} &\? \text{Per-Model Speedup Meas.} \\[3pt]
    \frac{\sum_{l = 1}^{L_{\text{dense}}}{\text{SoL Runtime}_l}}{\sum_{l = 1}^{L_{\text{dense}}}{\text{Meas. Runtime}_l}} &= \frac{\sum_{l = 1}^{L_{\text{sparse}}}{\text{SoL Runtime}_l}}{\sum_{l = 1}^{L_{\text{sparse}}}{\text{Meas. Runtime}_l}} \\[3pt]
    \text{Dense Per-Model \% of SoL} &= \text{Sparse Per-Model \% of SoL}
  \end{aligned}
\end{equation*}

At the aggregated per-model level, the SoL speedup is equal to the measured speedup when the sparse and dense models are equally optimized, such that the percentage of the per-model SoL latency for dense and sparse are equal.
\saurav{This section is a bit confusing to me. First, why is it called "validation", when we don't empirically validate the S.R? Second, what does "equally optimized" mean from a user's perspective?}



\section{Case Study}

\subsection{DL Practitioner}
\label{sec:case-study-dl-practitioner}

\cameron{These sparsity roofline figures for convnext and swin tiny need to be compacted somehow if we're going to show 4 of them. Also they still need to show the large block sizes.}

\begin{figure*}[!t]
    \centering
    \hfill%
    \begin{subfigure}[t]{0.46\textwidth}
        \includegraphics[width=\linewidth,valign=t]{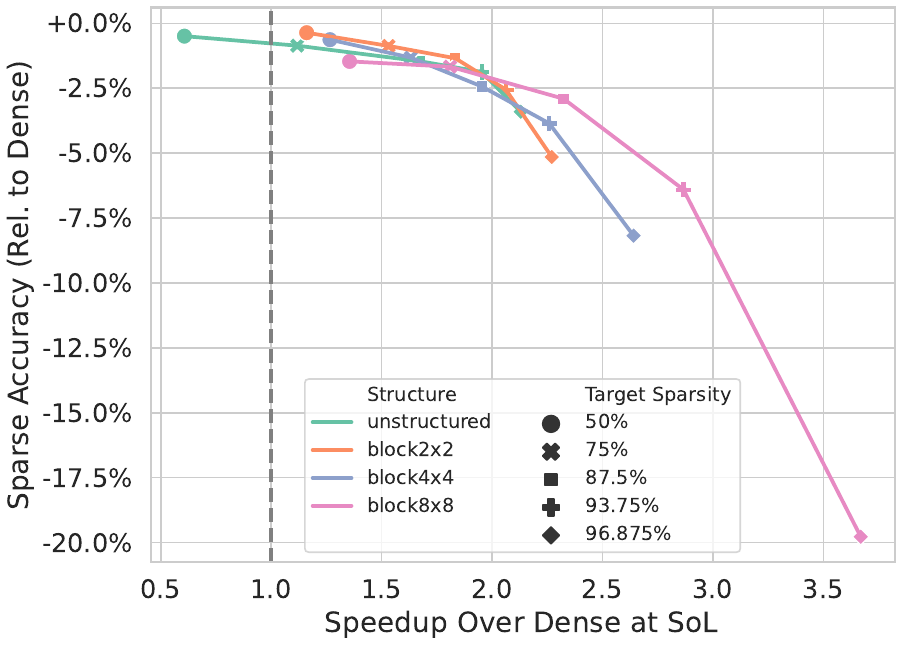}
        \caption{Block-Sparsity Roofline: ConvNeXt-Tiny on ImageNet-100}
        \label{fig:convnext_tiny_block_sparsity_roofline_b1}
    \end{subfigure}
    \hfill%
    \begin{subfigure}[t]{0.46\textwidth}
        \includegraphics[width=\linewidth,valign=t]{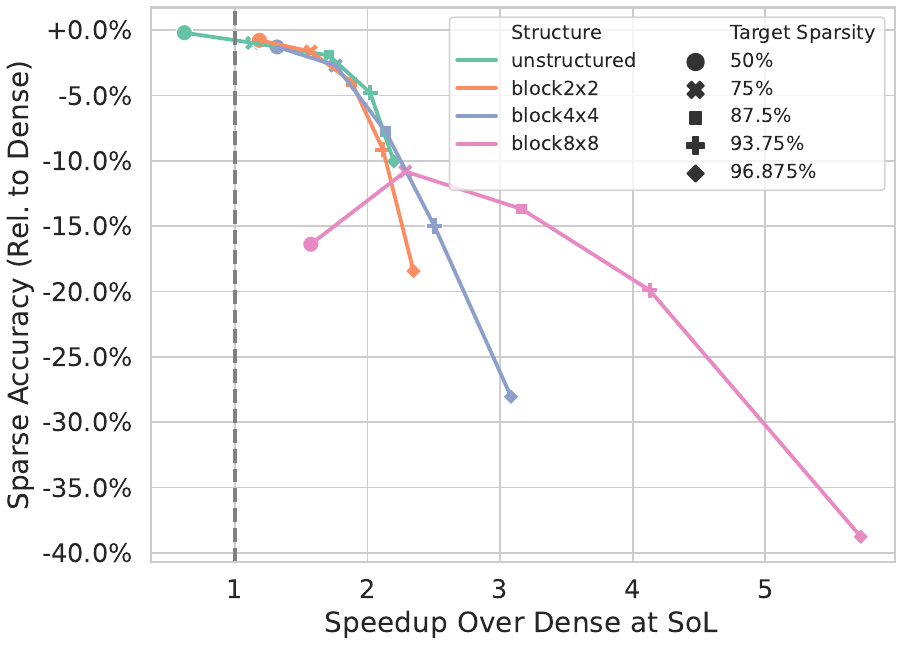}
        \caption{Block-Sparsity Roofline: Swin-Tiny on ImageNet-100}
        \label{fig:swin_tiny_block_sparsity_roofline_b1}
    \end{subfigure}
    \hfill%
    \caption{\textbf{Block-Sparsity Roofline}: The Sparsity Roofline for (a) ConvNext-Tiny and (b) Swin-Tiny on ImageNet-100 pruned with various block pruning sizes. Calculations are done using a batch size of 1 and NVIDIA A100 hardware specs.\collin{If we like the 2-col subfig, let's force the fig size the same in matplotlib (I tried to manually adjust it), or even better, do it in matplotlib so we can have a single legend on the bottom} \jowens{This is important: Use this caption to walk the reader through the conclusions he/she can make from looking at the SR\@. Tell don't show. The detail currently outlined in the body text is fine, but it is \emph{necessary} to describe here how the SR can be used to make the conclusions there.}}
    \label{fig:block_sparsity_roofline}
\end{figure*}

Suppose Alice is researching pruning algorithms and wants to find out whether block sparsity can provide effective inference latency improvements on NVIDIA GPUs for ConvNext~\cite{} and Swin~\cite{}, two state-of-the-art computer vision models.
She would typically start by training, pruning and then fine-tuning these models for various block sizes, say, 2$\times$2, 4$\times$4, 8$\times$8, 16$\times$16 and 32$\times$32, to capture a sufficiently large sample of the search space.

Alice would like to compare the speedups that her block pruning scheme achieves w.r.t.\ unstructured global magnitude pruning, but she would prefer to avoid implementing a custom block-sparse GPU kernel until she is sure it's the right approach.
She then considers using existing kernels from a vendor-optimized library such as cuSparse~\cite{}, but backs off due to two reasons: (1) writing a custom operator for a deep learning framework is not trivial, and (2) she notices in the documentation for the vendor-optimized library that it achieves poor performance for smaller block sizes, and may thus not provide a fair comparison across block sizes.

Rather than trying to measure actual latency numbers, Alice now plans to use some simple metrics to estimate potential speedups. She starts by counting the FLOPs of each sparse model.
However, since her blocked SpMM and unstructured SpMM kernels would be running on NVIDIA Tensor Cores and CUDA cores, respectively, the former will end up achieving higher throughput than the latter.
Additionally, since Tensor Cores necessitate more efficient memory bandwidth utilization, she would also need to account for the reads and writes that her sparse models perform during inference.

To address the above concerns, Alice instead generates the Sparsity Roofline for the block-sparse models she has trained to quickly approximate the speedups she would achieve for various block sizes.
Figures~\ref{fig:convnext_tiny_block_sparsity_roofline_b1} and~\ref{fig:swin_tiny_block_sparsity_roofline_b1} show the Sparsity Roofline models Alice would generate for ConvNext and Swin with a batch size of 1.
By observing the accuracy and performance tradeoffs that the Sparsity Roofline depicts, Alice is now able to determine that her models achieve higher speedups using larger block sizes, but they only maintain accuracy with smaller block sizes of 2$\times$2 and 4$\times$4. \emph{Importantly, Alice was able to arrive at this conclusion without needing to go to the effort of writing her own optimized sparse kernels for a variety of block sizes.} She now realizes that if she invests her time  in optimizing for smaller block sizes, she will get reasonable speedups without sacrificing accuracy.

\subsection{Hardware Architect}
\label{sec:case-study-hardware-architect}

\begin{figure*}[t]
    \centering
    \hfill%
    \begin{subfigure}[t]{0.46\textwidth}
        \includegraphics[width=\linewidth,valign=t]{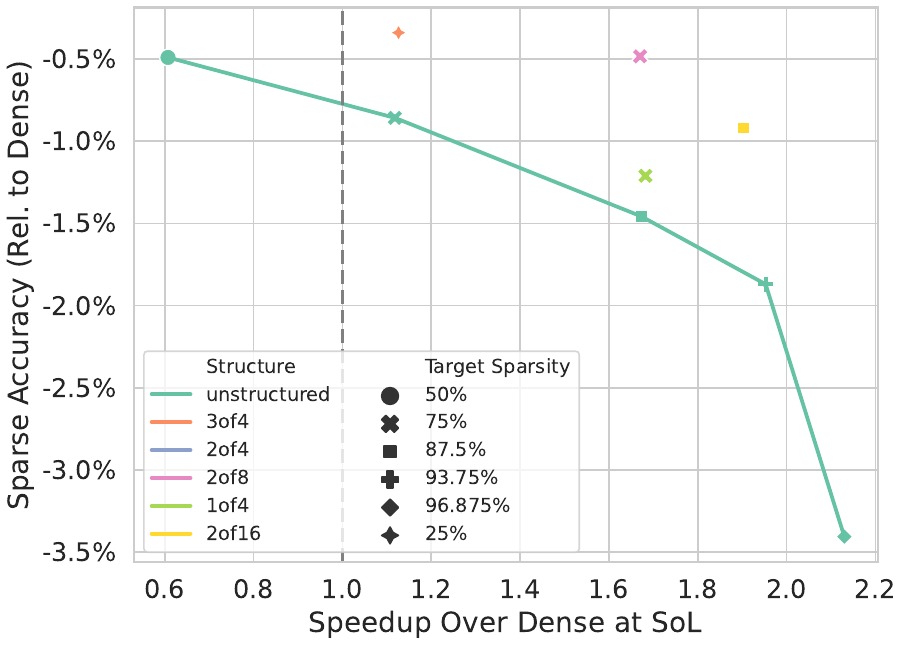}
        \caption{N:M Sparsity Roofline: ConvNeXt-Tiny on ImageNet-100}
        \label{fig:convnext_tiny_nofm_sparsity_roofline_b1}
    \end{subfigure}
    \hfill%
    \begin{subfigure}[t]{0.46\textwidth}
        \includegraphics[width=\linewidth,valign=t]{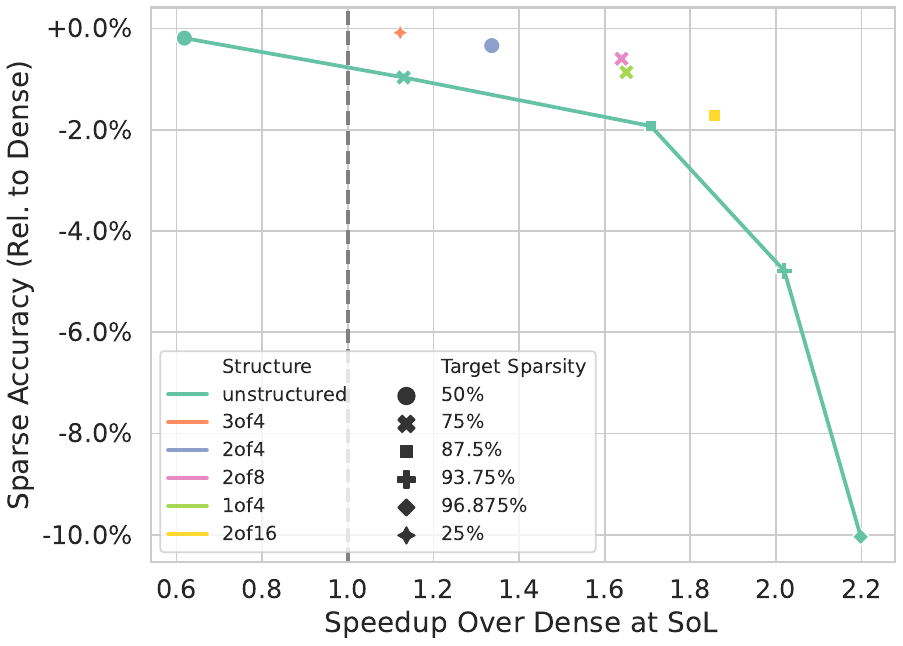}
        \caption{N:M Sparsity Roofline: Swin-Tiny on ImageNet-100}
        \label{fig:swin_tiny_nofm_sparsity_roofline_b1}
    \end{subfigure}
    \caption{\textbf{N:M Sparsity Roofline}: The Sparsity Roofline for (a) ConvNext-Tiny and (b) Swin-Tiny on ImageNet-100 pruned with various N:M patterns. Calculations are done using a batch size of 1 and NVIDIA A100 hardware specs.\collin{If we like the 2-col subfig, let's force the fig size the same in matplotlib (I tried to manually adjust it), or even better, do it in matplotlib so we can have a single legend on the bottom} \jowens{This is important: Use this caption to walk the reader through the conclusions he/she can make from looking at the SR\@. Tell don't show. The detail currently outlined in the body text is fine, but it is \emph{necessary} to describe here how the SR can be used to make the conclusions there.}}
    \label{fig:nofm_sparsity_roofline}
\end{figure*}

Bob is a hardware architect designing next-generation Tensor Cores for future GPUs and is investigating alternative N:M patterns for future hardware support.
He would like to quickly assess the accuracy and performance implications of the new N:M patterns before he puts in any effort into design and simulation.
His goal is to find patterns that achieve accuracy numbers similar to the currently supported 2:4 pattern, but are at least 30\% faster given the same Tensor Core throughput.

Bob's target workload for these N:M patterns is inference with a batch size of 1 on ConvNeXt and Swin. These two network architectures, in addition to providing state-of-the-art accuracies on their tasks, are also comprised of a variety of layer types, and involve matrix operations of various shapes and sizes, making them fairly representative.
The N:M schemes he chooses to investigate are 1:4, 2:8 and 2:16, in addition to the pre-existing 2:4 pattern.

Bob works with a machine learning engineer to get these two networks trained, pruned, and fine-tuned for each of the above sparsity patterns, and then obtains the corresponding accuracy numbers.
He now needs to determine how these models would perform if hardware support for the new N:M patterns was available.

Instead of developing RTL code for these new hardware units and simulating the workloads, which would be labor-intensive and time-consuming, Bob would prefer
a quicker way of estimating the runtime performance of each of these pruned models on their respective hypothetical hardware units.
Bob could simply use FLOPs to estimate speedups for each pattern (e.g., going from 2:4 to 1:4 is a 2x speedup); however, note that Bob would also need to account for the memory system's ability to keep up with the Tensor Core's throughput to get a more accurate performance estimation.

To address these concerns, Bob constructs the Sparsity Roofline for the N:M pruned models to quickly estimate the speedups he would achieve w.r.t.\ the accuracy. The resulting Sparsity Roofline plots are shown in Figures~\ref{fig:convnext_tiny_nofm_sparsity_roofline_b1} and~\ref{fig:swin_tiny_nofm_sparsity_roofline_b1}.
From the Sparsity Roofline, Bob notices that at the same Tensor Core throughput, 2:16 sparsity achieves nearly a 1.8$\times$ speedup over dense and is over 30\% faster than the 2:4 sparsity pattern, meeting his original goal.
He also notices that the 1:4 and 2:8 patterns are promising in cases where accuracy preservation is more important than raw speedup.
Similar to Alice (see Section~\ref{sec:case-study-dl-practitioner}), Bob was able to estimate his performance metrics significantly faster using the Sparsity Roofline.

\section{Discussion}
\label{sec:analysis}

\begin{figure}[tb]
  \centering
  \includegraphics[width=\linewidth,valign=t]{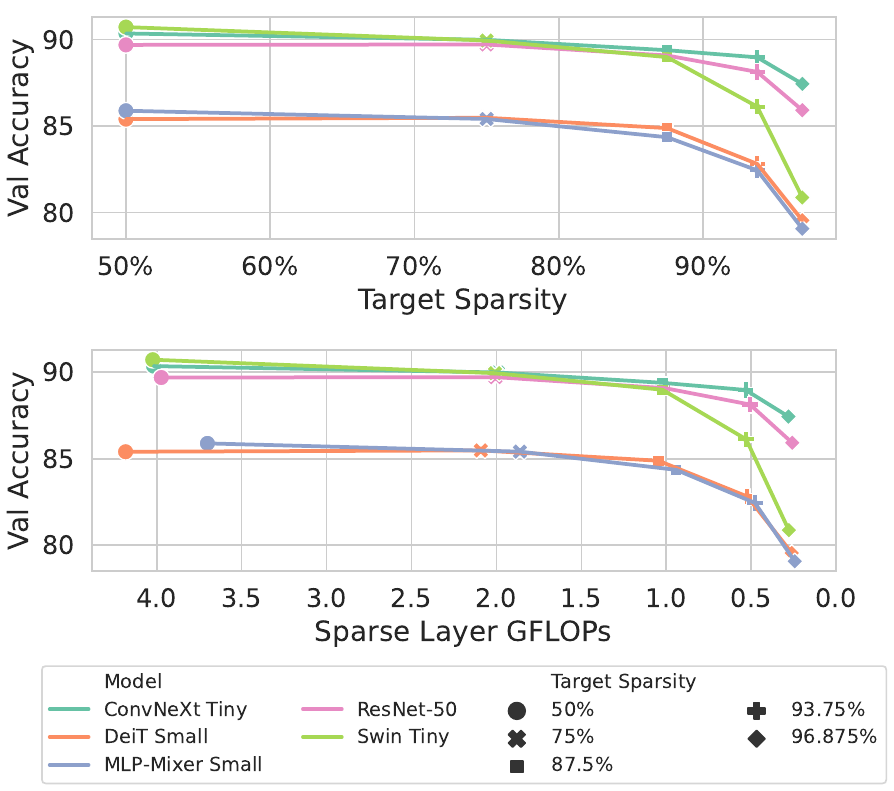}
  \caption{\textbf{Accuracy vs. Sparsity and FLOPs}: A common but misleading means of evaluating sparse models. Plotting accuracy (here ImageNet-100 top-1 accuracy) vs.\ sparsity (top) and FLOPs (bottom) for various models implies higher sparsity means higher GPU performance, which does not take memory bandwidth into account.\collin{I think sparsity needs to be ``normal'', from 0\% left to 100\% right. I describe it this way in the text below as well.}\collin{Do we want this as a sub-figure next to figure 3? Easy to do, but visually they're a bit different so I'm not sure} \mosama{This figure is not mentioned till Section~\ref{sec:analysis}! Just like Figure 1, it is chilling here with no ``purpose''. Folks will have to really scroll between the text back and forth to understand the content just because of the strict organization of figures.}}
  \label{fig:unstr_acc_vs_sparsity_and_flops_b1}
\end{figure}

\subsection{Unstructured Sparsity}

Global magnitude pruning with re-training \cameron{cite} has become a widely applicable technique due to its simplicity and effectiveness. Figure~\ref{fig:unstr_acc_vs_sparsity_and_flops_b1} shows how this technique  can reach almost 90\% sparsity with minimal accuracy loss.
In the context of small computer vision models, Figure~\ref{fig:unstr_sparsity_roofline_b1} indicates that accuracy can only be preserved to about a 1.5$\times$ speedup over dense.
While a 50\% speedup would be somewhat substantial, the time cost of fine-tuning may not be worthwhile in every scenario.
Additionally, a 50\% speedup is far less than what FLOP counts would suggest.
At 87.5\% sparsity, a network requires only $1/8$ the FLOPs of the original, yet Figure~\ref{fig:unstr_sparsity_roofline_b1} tells us that a 8$\times$ speedup is infeasible in any case.

To make sparsity generally viable from a performance perspective, we need to understand and alleviate the underlying factors that inhibit SpMM from achieving the speedups suggested by the FLOP reduction.
Despite the wide range of factors that affect SpMM kernel performance on GPUs, such as load balancing and efficient data reuse~\cite{Gale:2020:SGK, Bell:2009:ISM}, we only consider the factors that make up the Sparsity Roofline.
Thus, in our analysis, we account for FLOPs, bytes read/written, hardware peak throughput, and hardware peak memory bandwidth (the same as the Roofline model).

One of the most glaring downsides of unstructured sparsity is its inability to leverage the GPU's tensor cores that are effectively leveraged by dense models.\footnote{We note that sparse matrix tiling methods~\cite{Jiang:2020:AND, Hong:2019:AST, Zhaodong:2021:ETC} can effectively use tensor cores for unstructured SpMM. However, the Sparsity Roofline does not account for the specific sparse matrix pattern or any potential row/column reorderings, so we will not consider these in our analysis.}
The Roofline model in figure~\ref{fig:stacked_unstr_spmm_roofline} shows the elevated peak tensor core throughput above the peak CUDA core throughput.
For the A100, the tensor core throughput is 16x faster than the CUDA core throughput~\cite{NVIDIA:2020:NAT}.
To address the hardware discrepancy and put sparse and dense on a level playing field, we opt to investigate sparsity structures that can leverage the tensor cores.

\subsection{Block Sparsity}

  \begin{figure}
    \centering
    \includegraphics[width=\linewidth,valign=t]{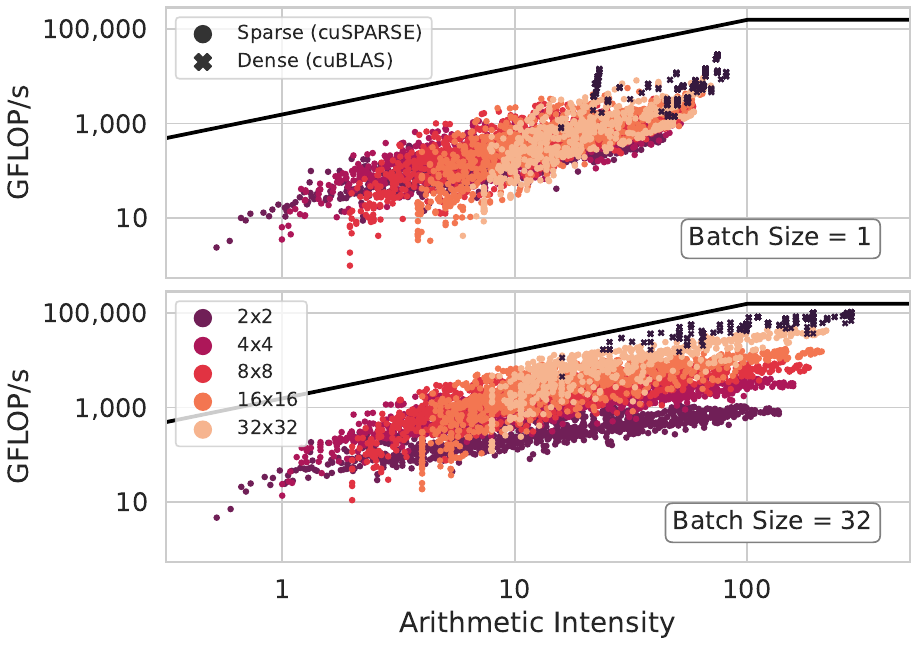}
    \caption{\textbf{Roofline, Block Sparse vs. Dense}: Roofline model measuring throughput of SpMM on all block sparse layers and GEMM on dense layers from all \jowens{replace ``all'' with a number} trained models, on a single NVIDIA A100.\collin{Should we call this ``operational intensity'' or should we add ``arithmetic intensity'' to related works when we introduce the roofline?} \mosama{This one should be where Fig~\ref{fig:unstr_acc_vs_sparsity_and_flops_b1} is.} \mosama{The caption explains nicely what I am looking at but what conclusion should I draw from a first glance? It could be as general as ``We can visualize the potential performance speedup\dots''} \jowens{AI should have units. I think properly the $y$ axis should be ``GFLOPs/s''.}}
    \label{fig:stacked_block_spmm_roofline}
  \end{figure}

The Sparsity Roofline shows two benefits of block sparsity: (1) the ability to use the high-throughput sparse tensor cores, and (2) the reduced index data from the block sparse format.
The reduced index data results from the sparsity pattern's more coarse-grained structure, where a single block index refers to multiple nonzeros.
The index data is reduced by a factor of the block size.

Despite the reduction in reads and writes from block sparsity, Figure~\ref{fig:stacked_block_spmm_roofline} shows that the vast majority of block-pruned weights are still memory bound.
Because of this, the Sparsity Rooflines for different block sizes in figure~\ref{fig:convnext_tiny_block_sparsity_roofline_b1} and figure~\ref{fig:swin_tiny_block_sparsity_roofline_b1} see only a small improvement compared to unstructured sparsity.
The accuracy-speedup tradoff is slightly better than unstructured sparsity at best, and only just as good in the worst case.

While the heatmap in Table~\ref{tab:structure_vs_unstructure} suggests that block sparsity should perform much better than unstructured, we observe that the accuracy loss from large block sizes (16$\times$16 and 32$\times$32) is too significant to be viable.
When we therefore restrict our analysis to smaller block sizes, we see that we can't achieve the full throughput from the tensor cores due to the memory bottleneck seen in Figure~\ref{fig:stacked_block_spmm_roofline}.
The smaller block sizes are completely memory-bound, whilst the larger block sizes are less so, and can thus get more throughput from the tensor cores.

\subsection{N:M Sparsity}

NVIDIA's sparse tensor cores provide an interesting alternative to block sparsity, allowing adopters to leverage the throughput of the tensor cores whilst being able to prune weights in a fine-grained manner.
While the coarse-grained structure of block sparsity restricts the freedom of pruning algorithms' weight selection and hurts accuracy, the fine-grained structured sparsity for the N:M patterns should theoretically hurt accuracy less.

In addition to the accuracy benefits of a fine-grained structure, the N:M formats can reduce the memory overhead for indexing data.
With dedicated hardware support, N:M formats only need $\log_2(M)$ bits to store the index of each nonzero inside the $M$-wide blocks; for 2:4, that's only 2 bits per nonzero.

Figures~\ref{fig:convnext_tiny_nofm_sparsity_roofline_b1} and~\ref{fig:swin_tiny_nofm_sparsity_roofline_b1} show the Sparsity Roofline for N:M formats. We see that the various N:M patterns achieve a better performance-accuracy tradeoff over unstructured than what block sparsity was able to achieve. N:M is an improvement over block sparsity in our pruned networks due to the reduced accuracy degradation and minimal index data overhead.

\begin{figure}
  \centering
  \includegraphics[width=\linewidth,valign=t]{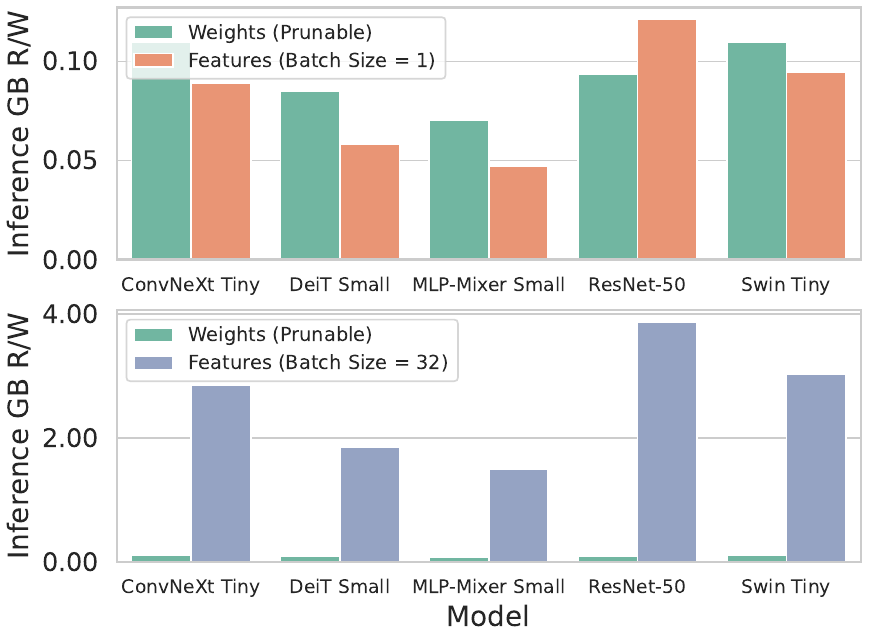}
  \caption{Memory accesses for weights versus input and output features, taken from dense models. The features take up a substantial portion of memory traffic that cannot be reduced via pruning.}
  \label{fig:small_weight_vs_feature_mem_access}
\end{figure}

\subsection{Feature Overhead}

Finally, we have not yet mentioned the read and write overhead of the input and output features of each layer.
Equation~\ref{eq:spmm_flops_gb} shows the data for the input and output features as $n \times k$ and $m \times n$ (respectively).
Akin to Amdahl's law, we can only expect to reduce the number of memory accesses for pruned matrices.
Therefore, regardless of our pruning strategy, the input and output features will always incur a fixed number of reads and writes as overhead.
Figure~\ref{fig:small_weight_vs_feature_mem_access} shows the severity of this problem.
For a batch size of 1, the feature memory accesses, which cannot be reduced via pruning, account for half of all accesses.
For a batch size of 32, the feature memory accesses heavily dominate the overall number of accesses, making it difficult to decrease the memory bottleneck of our sparse models.

The $n$ dimension in equation~\ref{eq:spmm_flops_gb} is shared by the input and output feature matrices and is not one of the weight matrix dimensions.
The size of $n$ relative to $m$ and $k$ is determines the appearance of the graphs in Figure~\ref{fig:small_weight_vs_feature_mem_access}.
The $n$ dimension scales linearly with both the batch size and the number of spatial locations in the feature data (for both convolution and transformer FFN layers).
This suggests that we will see larger speedups from pruning when the model size ($m$ and $k$) is large relative to the batch size and feature sizes ($n$).

\section{Related Work}

\paragraph{Automated Model Compression}
Recent work has explored various approaches for automatically inferring optimal sparsity levels using approaches such as Bayesian Optimization~\cite{Joseph:2020:APA} and reinforcement learning~\cite{He:2018:AAF}.
Our work differs in two ways: we focus on providing (1) a \emph{visual} representation of the accuracy and performance landscape for different sparsity patterns and levels, and (2) meaningful estimates of potential inference runtimes to aid deep learning practitioners, performance experts and hardware designers.

\paragraph{Deep Learning Roofline Models}
The Roofline model has been applied to the deep learning problem space in the past~\cite{Yang:2020:HRP, Wang:2020:TBR,Czaja:2020:ATR}.
However, this work primarily focuses on dense neural networks.
Specifically, \citet{Wang:2020:TBR} extend the Roofline model to deep learning by using latency and compute/bandwidth complexity.
\citet{Yang:2020:HRP} provide a toolkit extension for deep learning to support new precisions, tensor cores, and a tool for measuring performance metrics.
\citet{Czaja:2020:ATR} perform a Roofline analysis of DNNs accounting for non-uniform memory access (NUMA) systems.
\collin{Also we didn't address https://par.nsf.gov/servlets/purl/10077632 which is not the same as us but seems to be the closest thing}

\bibliography{sparsity}
\bibliographystyle{mlsys2023}

\clearpage
\appendix

\section{Appendix}

\subsection{Training Details}
\label{sec:training_details}

\begin{minipage}[t]{\linewidth}
    \resizebox{\textwidth}{!}{%
        \begin{tabular}{@{}ll@{}}
        \toprule
        Training Parameter           & Value           \\ \midrule
        Optimizer                    & AdamW           \\
        Base Learning Rate           & 1e-3            \\
        Weight Decay                 & 0.05            \\
        Optimizer Momentum           & $\beta_1,\beta_2=0.9,0.999$ \\
        Batch Size                   & 1024            \\
        Training Epochs              & 300             \\
        Learning Rate Schedule       & Cosine          \\
        Warmup Epochs                & 20              \\
        Warmup Schedule              & Linear          \\
        Warmup Learning Rate         & 0               \\
        Minimum Learning Rate        & 1e-5            \\
        Linear Scaling Learning Rate & False           \\
        Gradient Clip                & 5.0             \\ \bottomrule
        \end{tabular}
    }
    \captionof{table}{\textbf{Training Hyperparmeters}: Training hyperparameters used by all models.}
    \label{tab:train_hyperparams}
\end{minipage}

\begin{minipage}[t]{\linewidth}
    \resizebox{\textwidth}{!}{%
        \begin{tabular}{@{}ll@{}}
        \toprule
        Data Augmentation Parameter                         & Value                 \\ \midrule
        Random Augment~\cite{Cubuk:2020:RPA}                & Magnitude=9, Std=0.5  \\
        Random Augment Increasing~\cite{Wightman:2021:RSB}  & True                  \\
        Cutmix Probability~\cite{Yun:2019:CRS}              & 1.0                   \\
        Mixup Probability~\cite{Zhang:2018:MBE}             & 0.8                   \\
        Mixup Switch Probability                            & 0.5                   \\
        Mixup Mode                                          & Batch                 \\
        Random Erasing Probability~\cite{Zhong:2020:RED}    & 0.25                  \\
        Random Erasing Mode                                 & Pixel                 \\
        Random Erasing Count                                & 1                     \\
        Color Jitter                                        & 0.4                   \\
        Resizing Interpolation                              & Bicubic               \\
        Label Smoothing~\cite{Szegedy:2016:RIA}             & 5.0                   \\
        \bottomrule
        \end{tabular}
    }
    \captionof{table}{\textbf{Data Aug. Hyperparmeters}: Data augmentation hyperparameters used by all models via the \emph{timm} library~\cite{Wightman:2019:PIM}.}
    \label{tab:data_aug_hyperparams}
\end{minipage}

Table~\ref{tab:train_hyperparams} and Table~\ref{tab:data_aug_hyperparams} list the training and data augmentation hyperparameters used by all models. All data augmentations were performed using the \emph{timm} library~\cite{Wightman:2019:PIM}, which is also used by Swin~\cite{Liu:2021:STH}, DeiT~\cite{Touvron:2021:TDE} and ConvNeXt~\cite{Liu:2022:ACF}.

\subsection{Dataset}

We are releasing a sparse matrix dataset, containing the sparsity patterns of all pruned layers from all of the models we pruned. This dataset will allow performance experts to benchmark existing SpMM kernels and write new implementations that are optimized for the range of sparsity levels and structures that appear in deep learning. The matrices are organized by architecture, structure and sparsity as follows.

\begin{itemize}
    \item \textbf{Archictectures}: ResNet-50, EfficientNet-B4, ConvNeXt-Tiny, DeiT-Small, Swin-Tiny, MLP-Mixer-Small
    \item \textbf{Sparsity Patterns}: Block$2\times2$, Block$4\times4$, Block$8\times8$, Block$16\times16$, Block$32\times32$, Unstructured
    \item \textbf{Sparsity Levels}: 50\%, 75\%, 87.5\%, 93.75\%, 96.875\%
    \item \textbf{Matrices}: Global-magnitude pruned linear and convolution weight matrices
\end{itemize}

The resulting dataset contains 7,655 pruned matrices in the same matrix market file format used by the SuiteSparse Matrix Collection~\cite{Davis:2011:TUO}. The full dataset can be downloaded at:

\url{https://drive.google.com/file/d/1AItto4NeyiGRCYgtERMf0i0vflFt36Og/view?usp=sharing}

\end{document}